\documentclass[11pt,a4paper]{article}
\usepackage[utf8]{inputenc}
\usepackage[T1]{fontenc}
\usepackage{lmodern}
\usepackage{natbib}
\usepackage{graphicx}
\usepackage{amsmath}
\usepackage{booktabs}
\usepackage{multirow}
\usepackage{placeins}
\usepackage{hyperref}
\usepackage[margin=2.5cm]{geometry}

\title{Before the Word: Name Information in Description-Derived States of Small Language Models, With and Without Naming Failure}
\author{Fernando C\'ardenas Piepereit, Dipl.-Math. \\ V\'elez-M\'alaga, Spain}
\date{\today}

\begin{document}
\maketitle

\begin{abstract}
The tip-of-the-tongue (TOT) state, knowing an object but not its name, motivates this study as an analogy only. We ask whether small language models form a name-predictive state from a name-free description, and whether it differs when naming fails but description selection succeeds. In an internally pre-registered study (a frozen document, no external registry) of Qwen2-0.5B, SmolLM2-1.7B and Phi-3-mini with 200 items, we combine a similarity read-out, a Patchscopes-style read-out and activation patching with matched controls. H1 (name information in the state) was supported in all nine model-by-description-type combinations; the agreement rule was met by the similarity measure together with patching, while Patchscopes stayed near chance. Post-hoc robustness checks after internal review (logit lens, single-position patch, manipulation check) temper this reading: name information in middle layers is clearest under the multi-position patch and the similarity measure, and weak under single-position patching at 50\,\% depth. H2$'$ (description--paraphrase consistency of name-free states not lower in failure items) is mixed: the SmolLM2 deficit test is significant under the pre-registered specification but not robust; Phi-3 non-inferiority holds across margins and a word-overlap covariate, although its bootstrap interval includes zero; Qwen2-0.5B is inconclusive. A capacity explanation remains untested. We report vehicle-level signals only, not meaning.

\end{abstract}

\section{Introduction}\label{sec:intro}

A speaker can know exactly what an object is for and still fail to produce its name. In the classic tip-of-the-tongue (TOT) paradigm, participants are given a definition and asked for the word; in a subset of trials the word is not retrieved although the participant feels it is known and can often report partial properties of it \citep{brown1966tip,brown1991review}. In staged theories of lexical access, this pattern is explained by a separation between preparing a concept, selecting a lexical entry, and encoding its form \citep{levelt1999theory}. The motivation for the present study was a similar experience of the author: knowing what a technical object does without knowing a word for it. We use the human phenomenon only as motivation and as a declared analogy; we make no claim that the models share the experience or the underlying mechanism, and our measures do not operationalise the TOT state.

The question is the following. When a language model reads a functional description of an object, does it form an internal state that (i) is related to the object's name beyond what surface form and domain would predict, (ii) contributes causally to producing that name, and (iii) differs when the model fails to produce the name? We deliberately speak of \emph{name information} and of a \emph{vehicle-level signal above control}: we measure an activation state and compare it with matched controls. Following the vehicle/content distinction of an earlier study of activation states by the author \citep{cardenas2026architecture}, we report vehicle-level signals only and make no claim about meaning, content, or knowledge.

Related work probes parts of this question but, to our knowledge, not the combination. Reverse-dictionary prompts, which map a description to a word, have been used to analyse conceptual information in language models \citep{xu2024tip,xu2025revealing}. Names of entities can be read from intermediate states when the name itself is the input \citep{morand2025entities}. Intermediate entities that are never written out can be resolved internally \citep{yang2024latently,biran2024hopping}, and a model may hold more factual information internally than it produces \citep{gekhman2025insideout}. To our knowledge, none of these works compares a description-derived state with the name's own state layer by layer, tests it causally in a naming context, and splits items by whether naming succeeds (Section~\ref{sec:background}).

We therefore ask two questions that were fixed in an internal pre-registration, that is, a SHA-256-frozen document and not an entry in an external registry (Section~\ref{sec:prereg}). \textbf{H1:} the state after a description predicts the name, shown by a read-out criterion against matched control names and by a causal criterion (patching the state into a naming prompt), with the two required to agree. \textbf{H2$'$:} for items where the name is not produced but the right description is nevertheless selected (group A, ``production failure with description-selection success''), the description--paraphrase consistency of the name-free state is not lower than for items that are named correctly (group C). H2$'$ is thus an item-level comparison motivated by the TOT analogy; it measures consistency between states derived from a description and from its paraphrase, not the name. The stopping rule, the negative controls, and the correction for multiple tests were fixed before the final analysis; deviations are listed in Section~\ref{sec:methods} and Table~\ref{tab:deviations}.

We test them in three small open models, Qwen2-0.5B, SmolLM2-1.7B, and Phi-3-mini, using up to three description types per item (a freshly written description, a paraphrase, and, for 177 of 200 items, a dictionary gloss), which gives nine model $\times$ description-type combinations for H1. The negative controls include a randomly initialised model, descriptions of other items as donors, and hard negatives (semantically close items). After an internal review we added post-hoc robustness analyses and a pre-registered manipulation check that had been omitted from the first draft (Section~\ref{sec:res-posthoc}).

\paragraph{Results in brief.} H1 is supported in all nine combinations under the pre-registered rule. Patchscopes stayed near chance, so the agreement rule is met by the similarity measure together with patching, not by the two read-outs. The post-hoc checks qualify the picture: name information in middle layers is clearest under the multi-position patch, which approximates transplanting the description, and under the similarity measure; it is weak under single-position patching at 50\,\% depth. H2$'$ is mixed. The SmolLM2 deficit test is significant under the pre-registered specification but not robust, Phi-3-mini shows non-inferiority that survives alternative margins and a word-overlap covariate although its bootstrap interval includes zero, and Qwen2-0.5B yields a descriptive result only. Representational capacity is a candidate explanation of the model dependence; it is not tested here and is the central open question.

\paragraph{Contributions.}
\begin{itemize}
  \item An internally pre-registered pipeline that combines a similarity read-out, a Patchscopes-style read-out, and activation patching, with matched negative controls and an agreement rule.
  \item Evidence that, in three small models, a description-derived state in middle layers is predictive of the name (H1), together with post-hoc checks that show where this evidence is strong (similarity, multi-position patch) and where it is weak (single-position patch at 50\,\% depth, Patchscopes).
  \item A group-wise test (H2$'$) of whether the description--paraphrase consistency of name-free states differs when name production fails, with a model-dependent and not fully robust outcome that we report without smoothing.
  \item A statement of the capacity hypothesis as the central open question, and a documented list of protocol deviations.
\end{itemize}

\paragraph{Roadmap.} Section~\ref{sec:background} reviews lexical access, internal recall in language models, and probing and patching practice. Section~\ref{sec:methods} describes items, models, group formation, and measures. Sections~\ref{sec:results} and~\ref{sec:discussion} report and interpret the results, and Section~\ref{sec:limits} states limitations and future work.

\section{Background and related work}\label{sec:background}

\subsection{Lexical access and the tip-of-the-tongue state}

Models of speech production separate the preparation of a concept from the selection of a lexical entry and from the encoding of its form \citep{levelt1999theory}. The tip-of-the-tongue (TOT) state is usually read as a failure at a late step of such a sequence: the target is identified, but its word form is not fully available. The classic experimental route presents a definition and asks for the word; participants in a TOT state often report partial information about the word \citep{brown1966tip}. A review of the TOT experience is given by \citet{brown1991review}.

We use this literature only as motivation and for orientation; it is not an operational definition for our measures. A language model has no separate lexeme stage in the sense of staged production models, and we do not measure a feeling of knowing. The analogy therefore concerns a pattern of outcomes, namely failure to produce a name together with success in a related selection task, and not a shared mechanism. We return to the limits of the analogy in Section~\ref{sec:discussion}.

\subsection{Internal recall of entities and facts in language models}

Several lines of work show that information needed for an output can be present internally before, or without, being produced. Causal tracing was used to locate states that matter for factual associations in an autoregressive model \citep{meng2022locating}, and a later analysis dissected how attributes of a subject are extracted during factual recall \citep{geva2023dissecting}. For questions that require an unstated intermediate entity, \citet{yang2024latently} test whether the model resolves this entity internally. \citet{biran2024hopping} report that the intermediate entity is resolved in early layers while the second step happens late, and that patching a later-layer representation back into an earlier layer can correct up to 66\,\% of previously incorrect cases. \citet{gekhman2025insideout} find that a model can hold more factual knowledge internally than it expresses in its generations. These results motivate asking whether a failure to name an object coexists with an intact internal state. They concern factual questions and relational chains, not the retrieval of a name from a functional description. Activation patching between translation prompts has been used to argue that concept representations can be separated from the output language \citep{dumas2025separating}, and linear probes can recover numerical magnitudes from hidden states more reliably than models state them in their answers \citep{yuchi2026numbers}.

\subsection{From description to word}

The closest line of work uses the reverse-dictionary task, in which a model maps a description to a word. \citet{xu2024tip} use this task as a probe and report that internal representations encode object categories and finer features. Their title invokes the tip-of-the-tongue metaphor. Based on our reading of the full text, the study uses in-context demonstrations, does not patch activations or sweep layers, does not classify failures into types, and does not cite the TOT literature. A later study of the same group relates the same probe to human-like conceptual representations, including brain-activity data \citep{xu2025revealing}; in the part of it that we read, it likewise reads the representation from a single layer and does not intervene causally. \citet{morand2025entities} reconstruct entity names from hidden states with a layer sweep and report that intermediate layers work best and that frequency matters. Their input is the mention of the entity, not a description of it. We are not aware of a study that compares the state after a description with the state after the name, layer by layer, tests the description-derived state causally in a naming context, and splits items by naming success. This is a statement about the sources we found and read, and a search of this kind cannot exclude unseen work.

\subsection{Reading and testing internal states}

Our measures draw on three methodological traditions.

\paragraph{Probing and its pitfalls.} Supervised probes can detect information that the model does not use; control tasks and selectivity measures were proposed to separate the capacity of the probe from the content of the representation \citep{hewitt2019designing}, and a review discusses the limits of inferring use from decodability \citep{belinkov2022probing}. Probing methods can also disagree with one another \citep{bayazit2026lingua}. This is relevant here because our read-outs diverge (Section~\ref{sec:discussion}). For these reasons our read-out does not rely on a trained probe alone. We compare the description-derived state with the state of the target name against that of matched control names, using a similarity measure that corrects for hub-like neighbourhoods; this measure, CSLS, was introduced for word translation \citep{conneau2018word}.

\paragraph{Verbalising hidden states.} Patchscopes use the model itself to describe a hidden state by inserting it into a separate prompt, and include several earlier read-out methods as special cases \citep{ghandeharioun2024patchscopes}. We use this idea as a second read-out and report it separately.

\paragraph{Vehicle and content.} Reading an activation state is not the same as establishing what it means. Following the distinction between the vehicle (the activation state) and its content used in an earlier study of activation states by the author \citep{cardenas2026architecture}, the present paper reports vehicle-level signals only.

\paragraph{Activation patching.} Causal mediation analysis was applied to language models to ask which internal components mediate an effect \citep{vig2020causal}. Activation patching replaces an activation in one run with that from another run and measures the effect on the output. The results depend on design choices such as the corruption method and the metric, and recommendations for practice have been published \citep{zhang2024best,heimersheim2024activation}. We follow these recommendations where applicable and add controls that break the link between the patched state and the target (Section~\ref{sec:methods}).

\subsection{Capacity and shared representations}

If a model has fewer dimensions than features to represent, features may be stored in superposition \citep{elhage2022toy}. That analysis uses toy models, and its transfer to the models studied here is an assumption. We mention it because it provides one possible reason why a small model might show a deficit for rarely named items. We treat this as a hypothesis for the discussion, not as a result.

\subsection{Positioning}

In short, the literature supplies methods for reading names and entities from internal states, evidence that internal states can exceed outputs, and a reverse-dictionary probe for internal representations. What is missing, to our knowledge, is a design that (a) uses descriptions as input and the name's own state as the reference, (b) combines a read-out with a causal test and requires them to agree, and (c) asks, with matched controls, whether the description-derived state differs between naming failures and naming successes. The remainder of the paper addresses this combination in small models. It cannot speak to larger models, to the normal production pathway, or to meaning in a philosophical sense.

\section{Methods}\label{sec:methods}

The main analyses were specified in an internal pre-registration (version 1.2, SHA-256-frozen 2026-11-12; not deposited in an external registry) before any hidden-state measurement with the final prompt set. Section~\ref{sec:prereg} lists every departure from it, and Section~\ref{sec:posthoc} lists the analyses added after internal review. Throughout, a finding is a ``vehicle-level signal above control'' (name information in a state); we make no claim about meaning or content.

\subsection{Items}\label{sec:items}
The item set comprises 200 concrete nouns (``names''). Each item has a freshly written definition-style description (type i, primary), a paraphrase of it (type ii, reported as a replication), and, for 177 items, a WordNet gloss (type iii; \citealp{miller1995wordnet}), used for comparison only. The three types are never pooled. Each item also carries lists of synonyms and related words and one hard negative (a semantically close item with its own description). A further 30 minimal pairs cover 60 of the items. The items were written by language-model agents; the limitation is discussed in Section~\ref{sec:limitations}. Ten separate pilot items, not among the 200, were used for a patching pilot and for back-end verification.

\subsection{Models and back-ends}\label{sec:models}
We use three decoder-only models: Qwen2-0.5B \citep{yang2024qwen2} in fp32 on CPU, SmolLM2-1.7B \citep{allal2025smollm2} in fp16 on Apple MPS, and Phi-3-mini-128k-instruct \citep{abdin2024phi3} in 4-bit quantisation with MLX \citep{hannun2023mlx}. Phi-3 does not fit into 8\,GB of memory at full precision; its activations are computed in fp32 (Table~\ref{tab:deviations}). Qwen and SmolLM2 have $L=24$ layers, Phi-3 has $L=32$. A randomly initialised Qwen2-0.5B (seed 0) serves as null model for H1 only; the other families do not fit into memory.

Layer $l=1,\dots,L$ denotes the residual stream after block $l$; the embedding output is not tested, and all layers are measured. A depth fraction $f$ maps to layer $\mathrm{round\_half\_up}(f\,L)$. The \emph{middle range} is 40--75\,\% depth (layers 10--18 for $L=24$; 13--24 for $L=32$).

Reading and patching use a manual layer loop. Before any measurement it had to reproduce each model's own unpatched next-token log-probabilities on the 10 pilot items within a maximum absolute difference of 0.01. All four models passed (maximum difference 0.0; Phi-3 checked in its final fp32-activation configuration). An effect check confirmed that inserting another item's state at 50\,\% depth changes the output.

\subsection{Group formation}\label{sec:groups}
Groups were formed per model with a separate prompt set (G) that shares no few-shot example with the measurement prompts (set M, Section~\ref{sec:promptM}). Each item received a production outcome from greedy decoding and five samples ($T=0.8$) from a three-shot naming prompt, and a rank of its name (or, if better, of one of its listed synonyms) among all 200 names by sequence log-probability. A forced-choice \emph{description selection} test asked the model to choose the item's own description among the hard negative's description and two descriptions of items from other domains with the nearest word frequency. Options were scored by pointwise mutual information under a dictionary-entry template, averaged over four cyclic orderings, and the correct option additionally had to be more probable with the word than without it. Selection succeeded only if both criteria agreed on the correct option; disagreement placed the item in group U.

Groups are: C (name produced, with the name ranked in the top five), A (production failure with description-selection success), B (production failure without selection success), U (unclear; excluded) and X (unusable greedy output; excluded). Group sizes are given in Section~\ref{sec:results}. The stop rule of the project plan was triggered by these sizes. The planned comparison A versus B was therefore replaced by A versus C (H2$'$, Section~\ref{sec:h2}); this decision was made after the group sizes were known and before any hidden-state measurement under set M.

\subsection{Prompt set M}\label{sec:promptM}
All measurements use set M, which is disjoint from G: its few-shot examples (hammock, teapot, stethoscope) are neither items, pilot items nor part of G, and the words used in G are excluded. The descriptions $d$ are normalised to an initial capital and a final full stop. Names are always continuations with a leading space. The exact templates are given in Appendix~\ref{app:prompts}.

The primary H1 position is the last token (``called'') of NAMING\_M; the secondary position is the mean over the tokens of $d$. If they disagree, the primary position decides and the disagreement is reported.

\subsection{H1: measures of a name-predictive state}\label{sec:h1}
H1 states that the state derived from a description predicts the item's name (name information in the state). It is evaluated per model, description type and layer, with three measures. Let $s_d(l)$ be the state at the primary position of NAMING\_M and $s_n(l)$ the state at the last token of ``This object is called $\langle$name$\rangle$'' without few-shot prefix or description.

\paragraph{(a) Similarity.} For each item, the target name competes with 10 control names, drawn with seed $\mathrm{sha256}(\text{item id}+\text{model})$ from the 30 other items closest in model familiarity and with the same number of name tokens ($\pm1$ token if fewer than 30 exist; widened further for one SmolLM2 name, Table~\ref{tab:deviations}). The score is the reciprocal rank (rr) of the target among the 11 candidates by CSLS similarity \citep{conneau2018word} with $k=10$; neighbourhoods are taken over the 200 name and 200 description states without the item itself, and ties receive the mid-rank. Chance is $\mathrm{E}[\mathrm{rr}]=H_{11}/11\approx0.2745$. A baseline uses the same score on static input embeddings (mean over description tokens versus mean over name tokens). The measure holds in a layer if two tests pass at BH-adjusted $p<0.05$: a within-item label permutation of the target among the 11 candidates (5,000 permutations), and a one-sided paired sign-flip test of rr(model)$-$rr(baseline) (5,000 permutations). The larger $p$ enters the FDR family.

\paragraph{(b) Production read-out.} Following the Patchscopes approach \citep{ghandeharioun2024patchscopes}, $s_d(l)$ is inserted at the ``?'' position of the target prompt in layer $l$, and the next-token distribution is read. The score is the reciprocal rank of the target's first token among the first tokens of the target and the same 10 control names (a control sharing the target's first token is replaced by the next draw). Tests are as in (a), with the unpatched target prompt as baseline. No tuned lens is trained.

\paragraph{(c) Causal patching.} Following the activation-patching practice reviewed by \citet{zhang2024best,heimersheim2024activation} and the causal-intervention logic of \citet{vig2020causal,meng2022locating}, a donor state is written into a receiver, \texttt{FEWSHOT\_M + "Description: this object.\textbackslash nThis object is called"}, and the metric is $\log P(\text{full name}\mid\text{patched receiver})$, summed over name tokens (first-token $\log P$ is secondary). A \emph{single-position} patch replaces the receiver state at ``called'' in layer $l$. A \emph{multi-position} patch (middle range only) uses a receiver whose description consists of ``thing'' repeated to the donor's token count and copies the donor's description-token states onto the aligned positions; the receiver keeps its own few-shot prefix and template tail, so the effect is layer-dependent. In the middle range the multi-position patch is the primary patch type; elsewhere it is the single-position patch. Layers above 75\,\% depth are reported but never count as evidence.

\begin{table}[t]
\centering
\caption{Donors in the patching analysis. Gain for control $k$ is metric(a)$-$metric($k$).}\label{tab:donors}
\footnotesize
\begin{tabular}{@{}lp{0.80\textwidth}@{}}
\toprule
Donor & Definition \\
\midrule
(a) & The item's own description. \\
(b) & Descriptions of 5 other items (seed $\mathrm{sha256}(\text{item id})$); mean over the 5 (one receiver per donor length in the multi-position patch). \\
(c) & The hard negative's description. \\
(d) & The minimal-pair partner's description; only the 60 items in a minimal pair. \\
(e) & 5 random $\mathcal{N}(0,I)$ vectors (seed 20261112 + item index), each scaled to the norm of (a) at the same layer and position; mean over the 5. \\
(f) & Mean input embedding of the name tokens, scaled to the norm of (a) (per position in the multi-position patch). \\
(g) & Null model given the item's own description; Qwen only, single position only. \\
\bottomrule
\end{tabular}
\end{table}

For each control, a one-sided sign-flip permutation test over items (5,000 permutations) assesses a mean gain above zero (for (d), over its 60 items). The FDR family comprises layers $\times$ controls $\times$ patch types $\times$ models, separately per description type. A layer is \emph{(c)-significant} if, for its primary patch type, controls (b), (c) and (e) (and (g) for Qwen at single-position layers) are all BH-significant. Controls (d) and (f) are tested and reported separately (Table~\ref{tab:deviations}, D-M4-1); the literal rule that also requires (f) is reported alongside. The primary quantities are the \emph{emergence layer}, the smallest (c)-significant layer, and the gains at 50, 62.5 and 75\,\% depth.

\paragraph{Agreement rule.} H1 is supported for a model and description type if some layer at or below 75\,\% depth is (c)-significant and measure (a) or (b) holds, with both tests BH-significant, within $\pm2$ layers of it. Otherwise H1 is not supported. The measures need not agree: in our data Patchscopes (b) stay near chance, so the rule is met by the similarity measure (a) together with patching, and we report it as such. Disagreement between measures is reported with its direction and does not count as a finding. Because the multi-position patch copies all description states, the emergence layer within the middle range is bounded below by the switch of the primary patch type at 40\,\% depth and is partly a property of the design; single-position gains below 40\,\% are reported next to it.

\subsection{H2$'$: description--paraphrase consistency of name-free states when the name fails}\label{sec:h2}
H2$'$ asks whether, for items with production failure but description-selection success (group A), the name-free states of description and paraphrase are at least as consistent as for matched items the model names correctly (group C). Group C is restricted to items with description-selection success, which keeps the two groups symmetric.

\paragraph{Matching.} Pairs were formed from group-formation data only, blind to outcomes, by 1:1 nearest-neighbour matching without replacement on the Mahalanobis distance over five covariates (model familiarity, word frequency, name tokens, description words, $\log P$(description $\mid$ neutral template)), with calipers of 0.5\,SD on familiarity and on $\log P$(description). Matching starts with the least familiar A items. A third caliper on word frequency would have reduced the pairs below 30 and was not adopted; residual imbalance is handled by covariate adjustment. SmolLM2 and Phi-3 are confirmatory; Qwen, with fewer than 30 pairs, is descriptive only.

\paragraph{Name-free consistency.} States are read from READ\_M as the mean over the description tokens, for the description (type i, $s_i$) and the paraphrase (type ii, $s_{ii}$). Under causal attention these states cannot depend on any later naming prompt. The score is the CSLS rank (as in H1a) of an item's own $s_{ii}$ among all 200 $s_{ii}$ given $s_i$, with reciprocal ranks averaged over the middle range (40--75\,\% depth), fixed in advance independently of H1. As a sensitivity analysis, the first-token unembedding direction and the mean input embedding of the target name are projected out of both states. A lexical baseline computes the same score on static embeddings.

\paragraph{Estimation and tests.} For each pair, $\Delta=\mathrm{rr}(A)-\mathrm{rr}(C)$ is standardised by the SD of layer-averaged rr over all 200 items. The primary estimate is the intercept $b_0$ of an OLS regression of $\Delta$ on the five covariate differences (A$-$C, each standardised by its SD over the 200 items); the unadjusted mean $\Delta$ is secondary. This fits six parameters on 30 or 34 pairs, which we state as a limitation. Two one-sided sign-flip tests with 10,000 permutations are run per confirmatory model, flipping the sign of a random subset of pairs' outcomes and covariate differences and refitting the regression: non-inferiority (H0: adjusted $\Delta\le-0.30$, margin 0.30\,SD) and deficit (H0: adjusted $\Delta\ge0$). The four $p$-values (two models, two tests) form one BH family. Table~\ref{tab:verdict} defines the verdicts. The overall verdict uses the confirmatory models only: \emph{supported} (at least one support, no falsifier), \emph{mixed} (at least one of each), \emph{not supported} (no support, at least one falsifier), \emph{inconclusive} (neither). The primary test is repeated with the lexical-baseline score difference as an additional covariate; the primary analysis decides. The difference in H1 name measures between A and C serves only as a manipulation check (pre-registration \S5(b)), since C requires the name to be in the top five; it is reported in Section~\ref{sec:res-posthoc}. Rr values averaged over all 200 items (as opposed to over pairs) are used only to compare with the lexical baseline.

\begin{table}[t]
\centering
\caption{Verdict per confirmatory model in H2$'$ (BH-adjusted tests).}\label{tab:verdict}
\footnotesize
\begin{tabular}{@{}lll@{}}
\toprule
Non-inferiority & Deficit & Verdict \\
\midrule
significant & not significant & support \\
significant & significant & small deficit within the margin \\
not significant & significant & falsifier \\
not significant & not significant & inconclusive \\
\bottomrule
\end{tabular}
\end{table}

\subsection{Statistics}\label{sec:stats}
Multiple testing is controlled with the Benjamini--Hochberg procedure \citep{benjamini1995controlling} at $q=0.05$ within the families defined above; for H1(a) and (b) the family is layers $\times$ models per description type and measure. The resampling unit is the item (the pair in H2$'$). The analysis seed is 20261112 and the null model uses seed 0; derived seeds are stated with their uses. Type-iii analyses use the 177 items with a gloss. Numerical failures (NaN or inf) are excluded and counted. Exploratory analyses (A versus B on Qwen, hint tests, extended answer classes) receive no confirmatory correction.

\subsection{Pre-registration and deviations}\label{sec:prereg}
The pre-registration is an internal document, not an external registry entry; its integrity rests on the SHA-256 manifest. Before any measurement with set M, the pre-registration, the matching script and its output, the group-formation script and outputs, and all item files were frozen in a SHA-256 manifest (frozen 2026-11-12). Any later change to these files counts as a deviation. Table~\ref{tab:deviations} in Appendix~\ref{app:deviations} lists the deviations from the project plan, from the pre-registration text, and implementation details where the text left a choice open. Two deviations (D-M4-1, D-M4-2) were decided by the project lead; D-M4-1 was fixed after six smoke-test items and before any full-data analysis.

Further implementation choices: description spans are tokens overlapping the description string; donor (c) uses the hard negative's single description for all description types; donors (b) for types ii and iii are other items' descriptions of the same type; the pre-registered difference to donor (d) is complemented by a matched comparison using the item's own minimal-pair description. Random vectors in the multi-position patch are drawn per position (seed 20261112 + item index + 1000$(e+1)$), and null-model donor states are not rescaled. The project plan (v2.1) specified a TOST equivalence test for H2$'$; the pre-registration (v1.2) replaced it, before measurement, by a non-inferiority test plus a separate deficit test.

\subsection{Post-hoc analyses after internal review}\label{sec:posthoc}
After the internal review of the first draft, the following analyses were added on existing data (no new model runs). They are \emph{post hoc (after internal review), not pre-registered}, and are reported regardless of outcome (Section~\ref{sec:res-posthoc}). The only exception to this label is item 1.
\begin{enumerate}\setlength\itemsep{0pt}
\item A versus C on the H1 name measures (a)--(c) over the matched pairs, middle range; this is the manipulation check of pre-registration \S5(b), omitted in the first draft.
\item Single-position patch gains against controls (b) and (c) at 50, 62.5 and 75\,\% depth, to separate the single-position patch from the multi-position patch, which approximates transplanting the description.
\item Null-model donor (g) over layers 1--24 (Qwen).
\item Emergence layer with a minimum effect of 0.5 nats against controls (b), (c) and (e).
\item A permutation null for the maximum over layers of the mean rr of measure (a).
\item A logit-lens read-out of $s_d(l)$ (type i), scored by reciprocal rank as in (a), and a variant of (a) in which the first-token unembedding direction and mean input embedding of the target name are projected out of $s_d$.
\item H2$'$ robustness: bootstrap 95\,\% interval and leave-one-pair-out range of $b_0$, non-inferiority margins 0.2, 0.3 and 0.5, a lexical placebo (the H2$'$ score on static embeddings), and a word-overlap (Jaccard) covariate.
\item A string check for the item name and name stems in all descriptions, paraphrases and glosses.
\end{enumerate}

\section{Results}\label{sec:results}

All numbers in this section are taken from the single numbers file of the project (\texttt{results\_numbers.json}). Effect sizes are reported next to every significance statement. ``Depth'' denotes layer index divided by the number of layers $L$ (Qwen2-0.5B and SmolLM2-1.7B: $L=24$; Phi-3-mini: $L=32$). Gains are log-probability gains in nats; rr denotes reciprocal rank. Sections~\ref{sec:res-groups}--\ref{sec:res-h2} report the pre-registered analyses; Section~\ref{sec:res-posthoc} reports one pre-registered manipulation check and post-hoc robustness analyses.

\subsection{Group formation}\label{sec:res-groups}

Table~\ref{tab:groups} shows how the 200 items fell into the groups defined in Section~\ref{sec:groups} (C: name produced; A: production failure with description-selection success; B and U as defined there). The distribution differs strongly between models. Qwen2-0.5B produced the name for only 30 items (C) and fell into group A for 65 items; SmolLM2-1.7B and Phi-3-mini produced the name for 148 and 138 items, leaving 33 and 39 items in group A. Group B was small in all models (27, 5 and 6 items), and group U was large only for Qwen (78 items; 14 and 17 for SmolLM2 and Phi-3).

\begin{table}[ht]
\centering
\footnotesize
\caption{Group sizes per model (200 items each) and number of matched A--C pairs available for H2$'$.}
\label{tab:groups}
\begin{tabular}{lrrrrr}
\hline
Model & C & A & B & U & Matched A--C pairs \\
\hline
Qwen2-0.5B & 30 & 65 & 27 & 78 & 15 \\
SmolLM2-1.7B & 148 & 33 & 5 & 14 & 30 \\
Phi-3-mini & 138 & 39 & 6 & 17 & 34 \\
\hline
\end{tabular}
\end{table}

Matching of A items to C items yielded 15, 30 and 34 pairs for Qwen, SmolLM2 and Phi-3. The stop rule of the pre-registration led to the revised hypothesis H2$'$, with Qwen analysed descriptively and SmolLM2 and Phi-3 as confirmatory models (Section~\ref{sec:groups} and the deviation table).

\subsection{H1: description-derived states predict the name}\label{sec:res-h1}

H1 was supported in all nine model $\times$ item-type cells under the primary decision rule (control f outside the criterion; deviation D-M4-1). The literal frozen rule gives the same verdict in every cell. Table~\ref{tab:h1} lists, for each cell, the layers at which the read-out and causal criteria agreed (restricted to $\leq 75\,\%$ depth), the formal emergence layer, the maximal reciprocal rank (rr) of the similarity measure (a), the maximal rr of Patchscopes (b), and the maximal gain of the patched description state over the control that uses other descriptions (control b). The chance rr for measure (a) is 0.2745.

\begin{table}[ht]
\centering
\footnotesize
\caption{H1 results per item type and model. Agreement layers: layers (out of $L$) at which the pre-registered agreement rule holds, as a range. Emergence: first formally significant layer. rr(a): maximal reciprocal rank of the similarity measure (chance 0.2745). rr(b): maximal reciprocal rank of Patchscopes. Gain: maximal patching gain over control b at any layer (nats). All cells: H1 supported under both rules.}
\label{tab:h1}
\begin{tabular}{llcccccc}
\hline
Type & Model & $L$ & Agreement layers & Emergence & rr(a) & rr(b) & Gain (b) \\
\hline
i   & Qwen   & 24 & 14--18 & 10 & 0.760 & 0.287 & 7.76 \\
i   & SmolLM2 & 24 & 13--18 & 1  & 0.903 & 0.351 & 14.22 \\
i   & Phi-3  & 32 & 14--23 & 13 & 0.909 & 0.298 & 18.35 \\
ii  & Qwen   & 24 & 16--18 & 10 & 0.729 & 0.290 & 6.97 \\
ii  & SmolLM2 & 24 & 11--18 & 10 & 0.896 & 0.347 & 12.96 \\
ii  & Phi-3  & 32 & 14--23 & 13 & 0.886 & 0.300 & 16.61 \\
iii & Qwen   & 24 & 16--18 & 3  & 0.729 & 0.287 & 7.56 \\
iii & SmolLM2 & 24 & 14--18 & 3  & 0.879 & 0.341 & 12.85 \\
iii & Phi-3  & 32 & 16--24 & 13 & 0.874 & 0.300 & 16.24 \\
\hline
\end{tabular}
\end{table}

\paragraph{Middle-layer pattern.} In all cells the agreement layers lie between 11 and 24, in the middle of the networks (roughly 40--75\,\% depth), and none lies below layer 11. The maximal rr of measure (a) ranges from 0.729 to 0.760 in Qwen, 0.879 to 0.903 in SmolLM2 and 0.874 to 0.909 in Phi-3, against a chance level of 0.2745. Thus, in the middle layers, the state computed from the description (without the name in the context) is much closer to the state of the target name than to the states of ten matched control names. This is a name-predictive, item-level signal above control, as H1 requires; it does not by itself show how the model uses it. The maximal gains over other descriptions are 6.97--7.76 nats in Qwen, 12.85--14.22 in SmolLM2 and 16.24--18.35 in Phi-3. We do not interpret these differences as an effect of model size, because log-probability gains are not comparable across tokenisers and only three models are available.

The emergence of significance in the middle range is bounded by the design: the patching analysis switches to the multi-position variant at 40\,\% depth, so the emergence layers of 10 (Qwen, SmolLM2; $\approx 42\,\%$ of 24 layers) and 13 (Phi-3; $\approx 41\,\%$ of 32 layers) lie close to the layer at which this variant is evaluated. They are not a measured onset of the state, and we do not interpret them as the depth at which it arises (but see Section~\ref{sec:res-posthoc} for a post-hoc effect-size criterion).

\paragraph{Gains at fixed depths.} Table~\ref{tab:depth} (Appendix~\ref{app:depth}) gives the gains of the multi-position patch at 50, 62.5 and 75\,\% depth (type i, representative; the other types behave alike). Across all nine cells, the gain over other descriptions (control b) ranges from 6.79 to 16.61 nats at 50\,\% depth, from 6.03 to 10.64 nats at 62.5\,\% depth, and from 0.93 to 3.93 nats at 75\,\% depth. The gain over the hard-negative description (control c) is smaller: 0.83 to 3.86 nats at 50\,\%, and between $-0.03$ and 0.85 nats at 75\,\%. All gains in Table~\ref{tab:depth} have BH-adjusted $q = 0.0003$, except the control-c gain of Phi-3 at 75\,\% depth (discussed below). Gains over control b decrease with depth in every model, steeply between 62.5 and 75\,\%.

\paragraph{Patchscopes stay near chance.} The maximal rr of the Patchscopes read-out (b) is 0.287--0.290 in Qwen, 0.341--0.351 in SmolLM2 and 0.298--0.300 in Phi-3, compared with a chance level of 0.2745. Patchscopes thus stay close to chance in Qwen and Phi-3 and only modestly above it in SmolLM2. The read-out measures therefore do not agree with each other: the agreement rule is met by the similarity measure (a) together with the patching analysis, not by Patchscopes.

\paragraph{Early formal emergence layers carry no effect.} Three cells have formal emergence layers far below the middle range: SmolLM2 type i (layer 1) and Qwen and SmolLM2 type iii (layer 3). In layers 1--9 the absolute single-position gain against other descriptions (b) never exceeds 0.016 nats and against the hard negative (c) never exceeds 0.019 nats in these three cells (significance without effect size). The agreement layers of these cells (13--18, 16--18, 14--18) lie in the same middle range as in the other cells, so the verdicts do not depend on the early layers.

\paragraph{Phi-3 at 75\,\% depth: hard negative close to the own description.} In Phi-3 at layer 24 (75\,\% depth), the patch with the hard-negative description (control c) is, in types i and ii, almost as effective as the description of the item itself (type iii: 0.21 nats, significant). The gain over control c is 0.03 nats ($q = 0.4128$) in type i and $-0.03$ nats ($q = 0.8016$) in type ii, not distinguishable from zero; in type iii it is 0.21 nats ($q = 0.0042$), significant but small. This is consistent with category-level rather than item-level information at this depth, which we report as an observation. It does not change the verdict: the agreement layers of Phi-3 lie at or below layer 23 (types i and ii) and layer 24 (type iii), and the gain over control b remains significant ($q = 0.0003$) at layer 24 in all three types (1.11, 1.01 and 0.93 nats). In Qwen and SmolLM2, control c remains significant at 75\,\% depth in all item types (0.48 to 0.85 nats).

\subsection{H2$'$: description--paraphrase consistency of name-free states in group A versus group C}\label{sec:res-h2}

H2$'$ compares the description--paraphrase consistency of name-free states (READ\_M), measured as reciprocal rank, in matched pairs of group-A and group-C items. Non-inferiority of A relative to C (margin 0.30 SD) supports H2$'$; a significant deficit of A is a falsifier. Qwen enters descriptively only (15 pairs); SmolLM2 (30 pairs) and Phi-3 (34 pairs) are confirmatory. Table~\ref{tab:h2} gives the results. The overall verdict is \emph{mixed}: one confirmatory model supports H2$'$ and the other meets the pre-registered falsifier criterion, which, as shown below and in Section~\ref{sec:res-posthoc}, is not robust.

\begin{table}[ht]
\centering
\footnotesize
\caption{H2$'$ results per model. rr: layer-averaged reciprocal rank. $\Delta$: A$-$C difference in SD units, unadjusted; $b_0$: adjusted intercept (SD units). NI: non-inferiority test; Def: deficit test. $p$ unadjusted; $q$ BH-adjusted over the confirmatory tests (not defined for Qwen, which is descriptive). Verdicts follow the pre-registered rule.}
\label{tab:h2}
\resizebox{\textwidth}{!}{%
\begin{tabular}{llrrrrrrrrl}
\hline
Model & Role & Pairs & rr A & rr C & $\Delta$ & $b_0$ & $p_{\mathrm{NI}}$ ($q$) & $p_{\mathrm{Def}}$ ($q$) & Verdict \\
\hline
Qwen    & descriptive  & 15 & 0.797 & 0.749 & $+0.150$ & $-0.331$ & 0.5257 (--) & 0.2422 (--) & inconclusive \\
SmolLM2 & confirmatory & 30 & 0.777 & 0.933 & $-0.596$ & $-0.631$ & 0.8656 (0.8656) & 0.0181 (0.0362) & falsifier (not robust) \\
Phi-3   & confirmatory & 34 & 0.867 & 0.833 & $+0.148$ & $+0.267$ & 0.0118 (0.0362) & 0.8532 (0.8656) & support \\
\hline
\end{tabular}}
\end{table}

\paragraph{SmolLM2: deficit test significant, not robust.} In SmolLM2 the name-free states of group-A items are retrieved less well than those of matched group-C items (rr 0.777 versus 0.933). The adjusted intercept is $b_0=-0.631$ SD; non-inferiority is not shown ($p=0.8656$) and the deficit test is significant after correction ($p=0.0181$, $q=0.0362$). Under the pre-registered specification this is a falsifier of H2$'$ for this model. It is not robust, however: the bootstrap 95\,\% interval of $b_0$ includes zero (Section~\ref{sec:res-posthoc}), and the projection analysis below weakens it.

\paragraph{Phi-3: non-inferiority.} In Phi-3 the reciprocal ranks of A and C are similar (0.867 versus 0.833). The adjusted intercept is $b_0=+0.267$ SD, above the lower margin of $-0.30$ SD; non-inferiority holds ($p=0.0118$, $q=0.0362$) and no deficit is detected ($p=0.8532$, $q=0.8656$). The verdict is support. The confirmed non-inferiority concerns the margin and not equality; we make no claim that group A is better than group C.

\paragraph{Qwen: inconclusive.} For Qwen the unadjusted difference is positive ($+0.150$) while the adjusted intercept is negative ($-0.331$), and neither non-inferiority ($p=0.5257$) nor deficit ($p=0.2422$) is established with 15 pairs. We draw no conclusion for this model, and it does not enter the overall verdict.

\paragraph{Sensitivity analysis (pre-registered).} When the name directions are projected out of the states, the signs of $b_0$ stay the same in all three models (Qwen $-0.324$, SmolLM2 $-0.459$, Phi-3 $+0.349$). For Phi-3, non-inferiority holds ($p=0.0068$) and no deficit appears ($p=0.9125$). For SmolLM2, non-inferiority remains unsupported ($p=0.6966$), but the deficit test is no longer below 0.05 ($p=0.0734$, unadjusted). The point estimate stays negative ($-0.459$ SD), so this is a weakening, not a reversal, and it does not show that the group-A and group-C states are equivalent in this model.

\paragraph{Lexical covariate (pre-registered).} Adding the lexical-baseline score difference as covariate leaves the verdicts unchanged on unadjusted $p$-values. SmolLM2: $b_0=-0.706$, $p_{\mathrm{NI}}=0.9076$, $p_{\mathrm{Def}}=0.0103$. Phi-3: $b_0=+0.264$, $p_{\mathrm{NI}}=0.0164$, $p_{\mathrm{Def}}=0.8472$. Qwen: $b_0=-0.431$, $p_{\mathrm{NI}}=0.5926$, $p_{\mathrm{Def}}=0.1997$. BH-adjusted values for this analysis are not part of the numbers file. The lexical-baseline rr over all 200 items (0.289 for Qwen, 0.325 for SmolLM2, 0.257 for Phi-3) is well below the layer-averaged rr of the name-free states over all 200 items (0.696, 0.842 and 0.859).

\subsection{Pre-registered manipulation check and post-hoc robustness}\label{sec:res-posthoc}

The A-versus-C comparison of the H1 name measures is pre-registration \S5(b) (manipulation check) and was omitted in the first draft of this paper. All other analyses in this subsection are \emph{post hoc (after internal review), not pre-registered}; they were added in response to the internal review and are reported in full, whatever their direction. Methods are in Section~\ref{sec:posthoc}.

\paragraph{A versus C on the H1 name measures (pre-registered manipulation check).} Table~\ref{tab:ac} compares the matched pairs (15, 30, 34) in the middle range. If group A names fail because name information is weaker, A should score lower than C. Patching gains of A are lower than those of C against control b in all three models and against control c significantly only in Phi-3 (Qwen $p=0.0522$, SmolLM2 $p=0.0982$, unadjusted, two-sided). The similarity measure (a) does not differ significantly between A and C in any model; in Phi-3 the two are nearly equal (0.625 versus 0.605). We read this as reduced but not absent name information in group A.

\begin{table}[ht]
\centering
\footnotesize
\caption{Matched A--C pairs, middle range (type i). Entries are the A$-$C difference with unadjusted two-sided $p$ in parentheses. rr(a): similarity measure; gain (b), gain (c): multi-position patching gain in nats over other descriptions and over the hard negative.}
\label{tab:ac}
\begin{tabular}{lrrrr}
\hline
Model & Pairs & rr(a) & Gain (b) & Gain (c) \\
\hline
Qwen2-0.5B   & 15 & $-0.134$ (0.183) & $-2.10$ (0.010)  & $-0.86$ (0.052) \\
SmolLM2-1.7B & 30 & $-0.055$ (0.304) & $-2.47$ (0.0002) & $-0.78$ (0.098) \\
Phi-3-mini   & 34 & $+0.020$ (0.696) & $-1.52$ (0.004)  & $-1.07$ (0.022) \\
\hline
\end{tabular}
\end{table}

\paragraph{Single-position patch in the middle range (post hoc).} The large middle-range gains in Table~\ref{tab:depth} come from the multi-position patch, which copies all description-token states and so approximates transplanting the description itself; the informative comparison is control (c). The single-position patch at the last token is small at 50\,\% depth: for type i the gain against control b is 0.11 (Qwen), 0.06 (SmolLM2) and 0.15 nats (Phi-3), and against control c it is $-0.02$, $-0.01$ and $-0.20$ nats (not significant). It becomes substantial later: at 75\,\% depth against control b it is 4.58 (Qwen), 5.73 (SmolLM2) and 13.11 nats (Phi-3), against control c 0.70, 0.81 and 4.28 nats ($q=0.0003$). At 62.5\,\% depth only Phi-3 has a substantial single-position gain (5.55 nats; control c: 1.39), whereas Qwen and SmolLM2 have 0.28 and 0.50 nats with negative control-c gains. Type ii shows the same pattern; in type iii the control-c gains at 62.5\,\% depth are already positive (0.38 and 0.79 nats for Qwen and SmolLM2).

\paragraph{Null-model donor and effect-size emergence (post hoc).} For Qwen, the own state beats the state of the randomly initialised model in every layer 1--24 in all three types (gains between 3.41 and 13.65 nats; type i: 6.55 nats at 50\,\% depth, type ii 6.98, type iii 6.30). With a minimum effect of 0.5 nats against controls (b), (c) and (e), the emergence layer is 10 (Qwen), 10 (SmolLM2) and 13 (Phi-3) in all three types; this replaces the formal early layers of Table~\ref{tab:h1}.

\paragraph{Maximum-over-layers null for (a) (post hoc).} The observed maximum of the layer-wise mean rr (0.760, 0.903, 0.909) lies far above the 95\,\% quantile of the permutation null of the maximum (0.324, 0.325, 0.326; $p=0.0005$ for each model).

\paragraph{Logit lens and projection (post hoc).} A logit-lens read-out of the type-i state is near chance through most of the middle range and rises late (Table~\ref{tab:lens}). At 75\,\% depth its rr is 0.521 (Qwen), 0.393 (SmolLM2) and 0.617 (Phi-3); its maximum, at the last layer, is 0.742, 0.968 and 0.942. Similarity (a) exceeds it at every layer from about 50\,\% depth (layers 12--18 for $L=24$, 16--24 for Phi-3), so (a) is not merely the next-token read-out. Projecting the name directions out of $s_d$ barely changes (a): the maximum falls from 0.760 to 0.746 (Qwen), 0.903 to 0.896 (SmolLM2) and 0.909 to 0.892 (Phi-3); the middle-range mean changes from 0.409 to 0.423, 0.530 to 0.519 and 0.647 to 0.643.

\begin{table}[ht]
\centering
\footnotesize
\caption{Similarity (a) versus logit lens, type i, reciprocal rank at selected layers (chance 0.2745). Layers 12 and 18 for $L=24$; 16 and 24 for Phi-3 (50 and 75\,\% depth).}
\label{tab:lens}
\begin{tabular}{lrrrr}
\hline
Model & Layer & (a) & Logit lens \\
\hline
Qwen2-0.5B   & 12 & 0.362 & 0.276 \\
             & 18 & 0.679 & 0.521 \\
SmolLM2-1.7B & 12 & 0.392 & 0.288 \\
             & 18 & 0.820 & 0.393 \\
Phi-3-mini   & 16 & 0.424 & 0.297 \\
             & 24 & 0.899 & 0.617 \\
\hline
\end{tabular}
\end{table}

\paragraph{H2$'$ robustness (post hoc).} For SmolLM2 the bootstrap 95\,\% interval of $b_0$ is $[-1.398, 0.142]$, which includes zero; leave-one-pair-out values of $b_0$ range from $-0.774$ to $-0.481$. Non-inferiority is not supported at margins 0.2 and 0.5 either ($p=0.9284$ and $0.6691$). A lexical placebo (the same score on static embeddings) already yields $b_0=-0.384$ ($p_{\mathrm{Def}}=0.0996$), so part of the A--C difference is present before any contextual processing. With a word-overlap covariate (mean Jaccard overlap of description and paraphrase 0.036), $b_0=-0.687$ and $p_{\mathrm{Def}}=0.0166$. The SmolLM2 deficit is therefore significant under the pre-registered specification but not robust.

For Phi-3, $b_0=0.267$ has a bootstrap 95\,\% interval of $[-0.290, 0.747]$ (which includes zero), a leave-one-out range of 0.090 to 0.384, and non-inferiority holds at margins 0.2 and 0.5 ($p=0.0308$ and $0.0024$) and with the overlap covariate ($b_0=0.216$, $p_{\mathrm{NI}}=0.0252$). The lexical placebo gives $b_0=0.157$ ($p_{\mathrm{NI}}=0.0632$). For Qwen (descriptive), the bootstrap interval is $[-3.666, 2.896]$ and the leave-one-out range $-0.993$ to $0.578$; no margin yields support (smallest $p_{\mathrm{NI}}=0.3519$ at margin 0.5).

\paragraph{Leakage check (post hoc).} A string check found no item name in any freshly written description or paraphrase and no name stem in any text. One WordNet gloss (type iii) contains the name.

\paragraph{Summary.} H1 holds in all nine cells: description-derived states contain name-predictive, item-level information above control in the middle layers; the similarity measure and patching agree, Patchscopes do not. Group A shows reduced but not absent name information relative to group C. H2$'$ gives a mixed result under the pre-registered rule: group A is non-inferior to group C in Phi-3 (robust to margin 0.2--0.5 and to a word-overlap covariate), shows a deficit test significant but not robust in SmolLM2, and is inconclusive in Qwen. With two confirmatory models, the data do not support a statement about the dependence of this pattern on model size beyond the observation that it differs between models.

\section{Discussion}\label{sec:discussion}

\subsection{What H1 shows and what it does not}

In all nine model $\times$ description-type combinations, a state computed from a description, without the name in the context, lies closer to the name's own state than to the states of matched control names, and the multi-position patch of that state into a naming prompt has a causal effect in the same layers (Table~\ref{tab:h1}). The pre-registered agreement rule is met in the middle of the networks. We state precisely how: Patchscopes stayed close to chance, so the rule is met by the similarity measure (a) together with patching, not by independent agreement of all read-outs. The divergence between Patchscopes and the similarity measure is an instance of the broader observation that probing methods can disagree \citep{bayazit2026lingua}; we do not know which of them is closer to what the model uses.

The post-hoc checks (Section~\ref{sec:res-posthoc}) shape the reading. The large middle-layer gains come from the multi-position patch, which approximates transplanting the description itself, so gains over unrelated descriptions are close to guaranteed; the informative comparison is the hard-negative description (c), where gains are much smaller and, in Phi-3 at 75\,\% depth, indistinguishable from zero for types i and ii (type iii: 0.21 nats). The single-position patch at 50\,\% depth gives at most 0.15 nats against unrelated descriptions and at most 0.13 nats against (c) (negative in type i); it becomes substantial from 62.5\,\% (Phi-3) or 75\,\% depth (Qwen, SmolLM2). The logit lens is near chance through most of the middle range, and the similarity measure exceeds it at every middle layer from about 50\,\% depth, so the similarity signal is not only the next-token read-out. Projecting name directions out of the description state barely changes it. The manipulation check shows that patch gains against unrelated descriptions are lower in group A than in group C in all three models, while the similarity measure does not differ; name information is therefore reduced, but not absent, in group A. Together, name information is clearest under the multi-position patch and the similarity measure and is weak where only one position is transplanted at mid depth. At late depths in Phi-3 the state appears, in types i and ii, closer to the category than to the item. Here ``name-predictive'' means more informative about the name than the controls are, not ``encodes the name and nothing else''.

Further limits follow from the design. Causal sufficiency in a constructed context is not the normal pathway: it shows that the state \emph{can} drive the name there, not that the model uses it when naming an object. The middle-layer location states where our measures hold, and emergence layers are partly fixed by the switch of the primary patch type at 40\,\% depth; with a minimum effect of 0.5 nats the emergence layers are 10, 10 and 13 in Qwen, SmolLM2 and Phi-3. We make no claim about meaning or understanding, in line with the vehicle/content distinction \citep{cardenas2026architecture}. Decodability alone does not show use \citep{hewitt2019designing,belinkov2022probing}, and patching outcomes depend on design choices \citep{zhang2024best,heimersheim2024activation}.

\subsection{H2$'$ is mixed, model-dependent, and not fully robust}

The measure in H2$'$ is the description--paraphrase consistency of name-free states, not a measure of the name. In Phi-3, the states of items that the model fails to name are non-inferior to those of matched named items. This holds for margins of 0.2, 0.3 and 0.5\,SD ($p=0.031$, $0.016$, $0.002$) and with word overlap as a covariate ($p=0.025$; mean Jaccard overlap 0.036), but the bootstrap 95\,\% interval of the adjusted intercept ($0.267$\,SD) is $[-0.29, 0.75]$ and includes zero. In SmolLM2, the deficit test is significant under the pre-registered specification ($q=0.036$), but not robust: the bootstrap interval of the intercept ($-0.631$) is $[-1.40, 0.14]$, the projection sensitivity gives $p=0.073$, and a lexical placebo shows that part of the A--C difference is already present in static embeddings ($p=0.10$). We therefore do not describe the SmolLM2 result as a falsification. In Qwen2-0.5B, too few pairs were available for a confirmatory test and the result is inconclusive. The overall verdict remains \emph{mixed}; two confirmatory models, 30 and 34 pairs, and six regression parameters do not establish how often either pattern occurs, and the Phi-3 result does not show that the groups are equal. Group A is defined by the model's own outputs, so the groups differ in other respects that matching reduces but cannot remove.

\subsection{A capacity reading, stated as a conjecture}

One candidate explanation for the difference between SmolLM2 and Phi-3 is representational capacity. If a model has fewer dimensions than features to represent, features can be stored in overlapping directions \citep{elhage2022toy}. A small model might then show a less distinct description-derived state when a name is weakly available, while a larger model could keep the two apart. This is a conjecture. The source analysis concerns toy models; the two confirmatory models differ in size, but also in architecture, training data, precision and, for Phi-3, in 4-bit weights. Because the SmolLM2 deficit is itself not robust, the pattern to be explained is weaker than first reported. The projection sensitivity is compatible with the reading but does not test it, as it removes only two directions and crosses the 0.05 line of an unadjusted test. We do not read the H1 gains across models as a size trend: values in nats are not comparable across tokenisers. Size-dependent changes can look abrupt or smooth depending on the metric \citep{wei2022emergent,schaeffer2023mirage}. Section~\ref{sec:limitations} sets out a test.

\subsection{The limits of the human analogy}

The TOT literature motivated the question \citep{brown1966tip,brown1991review,levelt1999theory}, and the analogy should not be pushed. A language model has no separate lexeme stage; we do not measure a feeling of knowing, central to the human state; and we have not tested partial access to word form such as initial letters (Section~\ref{sec:limitations}). Group A is defined operationally by production failure and successful selection of the matching description. In Qwen, group A has 65 items but only 30 items are named correctly (group C), so format and ability floors make A there no TOT analogue. Whether any pattern amounts to a blockade between a preserved state and its form cannot be decided from state consistency alone. The analogy concerns a pattern of outcomes and nothing else.

\subsection{Width and test-time reasoning}

If capacity matters, it is capacity in width and depth, the pre-verbal space our method measures. This differs from test-time reasoning, which unfolds in words and passes through routes our method cannot reach. Closed frontier models are outside this method, since it requires access to internal states.

\subsection{Summary of the interpretation}

The data support one statement with moderate confidence: in three small open models, description-derived states in the middle layers are predictive of the name beyond matched controls, most clearly under the similarity measure and the multi-position patch. They do not support a general statement about what happens when naming fails: the H2$'$ outcomes are model-dependent and, for SmolLM2, not robust. The capacity reading is a hypothesis for future tests.

\section{Limitations and future work}\label{sec:limitations}\label{sec:limits}

\subsection{The capacity hypothesis: the central open question}\label{sec:capacity}

The main open question is whether the separation between a description-derived state and the name's own state depends on model size. The hypothesis, stated as a conjecture in Section~\ref{sec:discussion}, is that small models represent an object and its name in strongly overlapping directions, so that a missing name goes together with a less distinct state, whereas larger models can keep the two apart. Our evidence is two confirmatory models with different H2$'$ verdicts, one of which (SmolLM2) is not robust, and a sensitivity analysis that weakens it. Size is confounded with model family, training data and precision, so none of this tests the hypothesis. We do not use differences in H1 gains between models as evidence, because values in nats are not comparable across tokenisers.

A test would apply the same pipeline to a series of open-weight models, ideally within one family, for instance at roughly 7--8B, 13--14B and 70B parameters. The prediction is that the A$-$C difference in H2$'$ increases, or the deficit shrinks, with size. Because size-dependent changes can look abrupt or smooth depending on the metric \citep{wei2022emergent,schaeffer2023mirage}, the series should be analysed with more than one metric, and the prediction should be pre-registered before the runs. Models of 13B parameters and above do not fit on the 8\,GB machine used here (a 7B model ran in 4-bit in an earlier project, but slowly), so a full series requires GPU access; no such access is planned at the time of writing. Test-time reasoning in words is out of reach of our method, and closed frontier models cannot be examined at all.

\subsection{Analyses not run}

The pre-registration lists exploratory analyses that were not performed, by decision of the project lead after the confirmatory results were in: a comparison of groups A and B in Qwen2-0.5B (27 items in B) with the H1 measures and description--paraphrase consistency; hint tests on groups A and C (initial letter, first syllable, related word, with uninformative, wrong-letter and random-letter controls); extended answer classes (WordNet distance of at most three edges as ``wrong but related''); probes for initial letter and syllable count with control tasks and an n-gram baseline; and a reverse-direction patch from the name state into a ``describe'' prompt. In particular, nothing in this paper bears on partial access to word form, which is the signature of the human TOT state. Two further checks suggested by the internal review were not run because they need new runs: re-forming the Phi-3 groups in fp32, and control inputs with scrambled descriptions or word lists in place of descriptions.

\subsection{Further limitations}

\paragraph{Models and precision.} Only three small models were studied, and none is a model for which the conclusions would be expected to transfer without further tests. Phi-3-mini ran with 4-bit weights; its activations were computed in fp32 after the cache check failed in fp16 (D-M4-2), but its group formation ran in fp16 and was not repeated (Table~\ref{tab:deviations}). Fp16 noise may flip borderline group assignments, and differences between Phi-3 and the other models may include quantisation and precision effects.

\paragraph{Weak read-out and the multi-position patch.} The Patchscopes read-out stayed near chance in all models, so the agreement rule is met by the similarity measure together with patching. The multi-position patch approximates transplanting the description, and the single-position patch at 50\,\% depth shows little name information (Section~\ref{sec:res-posthoc}). The primary patch type switches at 40\,\% depth, which bounds emergence layers from below; we do not interpret them as onsets.

\paragraph{Post-hoc analyses.} The robustness checks of Section~\ref{sec:res-posthoc} were added after internal review, and are not pre-registered; the exception is the comparison of groups A and C on the H1 name measures, which is the pre-registered manipulation check and was omitted from the first draft. Group A retains reduced but non-zero name information, so the contrast A versus C in H2$'$ is not a contrast of name-free versus name-bearing states.

\paragraph{Internal pre-registration and H2$'$.} The pre-registration is an internal, SHA-256-frozen document, not an external registry entry, and the plan version 2.1 specified an equivalence test that the pre-registration replaced by a non-inferiority test plus a separate deficit test before measurement (Table~\ref{tab:deviations}). The H2$'$ regression fits six parameters on 30 or 34 pairs, Qwen with 15 pairs is descriptive only, and the 0.30\,SD margin is not independently justified (results for 0.2 and 0.5\,SD are reported). H2$'$ replaced the original comparison after the group sizes were known, although before any hidden-state measurement with the final prompt set. Residual imbalance in word frequency is handled only by adjustment, and the lexical placebo shows that part of the A--C difference in SmolLM2 is present already in static embeddings.

\paragraph{Items and leakage.} The 200 items and their descriptions were written by language-model agents. Their register may differ from human-written definitions, and the number of items limits the subgroups that can be analysed. The WordNet glosses (type iii) partly mitigate this but cover 177 items; one gloss contains the name, and no fresh description or paraphrase does. In Qwen2-0.5B only 30 items are named correctly, so group C is small and a format or ability floor is possible.

\paragraph{Secondary sources for the classic TOT studies.} The characterisation of the classic tip-of-the-tongue procedure and of the findings attributed to Brown and McNeill (1966) and Brown (1991) rests on secondary summaries, not on the original articles; this affects only the motivation, not any result.

\paragraph{Scope of claims.} All findings concern vehicle-level signals compared with controls in single-pass states \citep{cardenas2026architecture}. They do not show how the model names objects in ordinary use, nor do they bear on meaning.

\section*{Data and code availability, and AI declaration}\label{sec:availability}

This work was conducted with the assistance of AI agents (Claude, Anthropic) operating under a defined, versioned research methodology (structured project phases, pre-registered success criteria fixed before each empirical test, mandatory negative controls / null baselines, and critical review simulated by the same model prompted to argue from explicitly divergent theoretical positions; these simulated reviewer roles are not independent human reviewers). Factual claims and citations were checked against their sources by AI agents (not independently of the drafting tool) prior to inclusion; two items could not be verified against primary sources (the descriptions of the classic tip-of-the-tongue procedure and findings attributed to Brown and McNeill (1966) and Brown (1991), which were checked against secondary summaries only) and are listed under Limitations. Findings are reported including null and negative results.

The human author set the research question and the scope and made all methodological decisions, including the deviations listed in Table~\ref{tab:deviations}. The AI agents performed the literature search, wrote the items and code, ran the experiments and analyses, and drafted the text. The items, scripts, the pre-registration with its SHA-256 freeze manifest, and the raw results are available on Zenodo (\href{https://doi.org/10.5281/zenodo.24228887}{doi:10.5281/zenodo.24228887}).

\bibliographystyle{plainnat}
\bibliography{references}

\begin{thebibliography}{31}
\providecommand{\natexlab}[1]{#1}
\providecommand{\url}[1]{\texttt{#1}}
\expandafter\ifx\csname urlstyle\endcsname\relax
  \providecommand{\doi}[1]{doi: #1}\else
  \providecommand{\doi}{doi: \begingroup \urlstyle{rm}\Url}\fi

\bibitem[Abdin et~al.(2024)]{abdin2024phi3}
Marah Abdin et~al.
\newblock Phi-3 technical report: A highly capable language model locally on
  your phone, 2024.

\bibitem[Bayazit et~al.(2026)Bayazit, AlKhamissi, and
  Bosselut]{bayazit2026lingua}
Deniz Bayazit, Badr AlKhamissi, and Antoine Bosselut.
\newblock Lingua franca or probing artifact? {R}ethinking latent language in
  multilingual {LLMs}, 2026.

\bibitem[Belinkov(2022)]{belinkov2022probing}
Yonatan Belinkov.
\newblock Probing classifiers: Promises, shortcomings, and advances.
\newblock \emph{Computational Linguistics}, 48\penalty0 (1):\penalty0 207--219,
  2022.
\newblock \doi{10.1162/coli_a_00422}.

\bibitem[Ben~Allal et~al.(2025)Ben~Allal, Lozhkov, Bakouch, Blázquez, Penedo,
  Tunstall, Marafioti, Kydlíček, Piqueres~Lajarín, Srivastav, Lochner,
  Fahlgren, Nguyen, Fourrier, Burtenshaw, Larcher, Zhao, Zakka, Morlon, Raffel,
  von Werra, and Wolf]{allal2025smollm2}
Loubna Ben~Allal, Anton Lozhkov, Elie Bakouch, Gabriel~Martín Blázquez,
  Guilherme Penedo, Lewis Tunstall, Andrés Marafioti, Hynek Kydlíček,
  Agustín Piqueres~Lajarín, Vaibhav Srivastav, Joshua Lochner, Caleb
  Fahlgren, Xuan-Son Nguyen, Clémentine Fourrier, Ben Burtenshaw, Hugo
  Larcher, Haojun Zhao, Cyril Zakka, Mathieu Morlon, Colin Raffel, Leandro von
  Werra, and Thomas Wolf.
\newblock {SmolLM2}: When smol goes big -- data-centric training of a small
  language model, 2025.

\bibitem[Benjamini and Hochberg(1995)]{benjamini1995controlling}
Yoav Benjamini and Yosef Hochberg.
\newblock Controlling the false discovery rate: A practical and powerful
  approach to multiple testing.
\newblock \emph{Journal of the Royal Statistical Society: Series B
  (Methodological)}, 57\penalty0 (1):\penalty0 289--300, 1995.
\newblock \doi{10.1111/j.2517-6161.1995.tb02031.x}.

\bibitem[Biran et~al.(2024)Biran, Gottesman, Yang, Geva, and
  Globerson]{biran2024hopping}
Eden Biran, Daniela Gottesman, Sohee Yang, Mor Geva, and Amir Globerson.
\newblock Hopping too late: Exploring the limitations of large language models
  on multi-hop queries.
\newblock In \emph{Proceedings of the 2024 Conference on Empirical Methods in
  Natural Language Processing (EMNLP 2024)}, 2024.
\newblock \doi{10.48550/arXiv.2406.12775}.

\bibitem[Brown(1991)]{brown1991review}
Alan~S. Brown.
\newblock A review of the tip-of-the-tongue experience.
\newblock \emph{Psychological Bulletin}, 109\penalty0 (2):\penalty0 204--223,
  1991.
\newblock \doi{10.1037/0033-2909.109.2.204}.

\bibitem[Brown and McNeill(1966)]{brown1966tip}
Roger Brown and David McNeill.
\newblock The ``tip of the tongue'' phenomenon.
\newblock \emph{Journal of Verbal Learning and Verbal Behavior}, 5\penalty0
  (4):\penalty0 325--337, 1966.
\newblock \doi{10.1016/S0022-5371(66)80040-3}.

\bibitem[C{\'a}rdenas~Piepereit(2026)]{cardenas2026architecture}
Fernando C{\'a}rdenas~Piepereit.
\newblock Architecture-dependent causal transfer of activation states across
  large language models, 2026.
\newblock Version 2.

\bibitem[Conneau et~al.(2018)Conneau, Lample, Ranzato, Denoyer, and
  J{\'e}gou]{conneau2018word}
Alexis Conneau, Guillaume Lample, Marc'Aurelio Ranzato, Ludovic Denoyer, and
  Herv{\'e} J{\'e}gou.
\newblock Word translation without parallel data.
\newblock In \emph{International Conference on Learning Representations (ICLR
  2018)}, 2018.
\newblock \doi{10.48550/arXiv.1710.04087}.

\bibitem[Dumas et~al.(2025)Dumas, Wendler, Veselovsky, Monea, and
  West]{dumas2025separating}
Cl{\'e}ment Dumas, Chris Wendler, Veniamin Veselovsky, Giovanni Monea, and
  Robert West.
\newblock Separating tongue from thought: Activation patching reveals
  language-agnostic concept representations in transformers.
\newblock In \emph{Proceedings of the Annual Meeting of the Association for
  Computational Linguistics (ACL 2025)}, 2025.
\newblock URL \url{https://aclanthology.org/2025.acl-long.1536}.

\bibitem[Elhage et~al.(2022)Elhage, Hume, Olsson, Schiefer, Henighan, Kravec,
  Hatfield-Dodds, Lasenby, Drain, Chen, Grosse, McCandlish, Kaplan, Amodei,
  Wattenberg, and Olah]{elhage2022toy}
Nelson Elhage, Tristan Hume, Catherine Olsson, Nicholas Schiefer, Tom Henighan,
  Shauna Kravec, Zac Hatfield-Dodds, Robert Lasenby, Dawn Drain, Carol Chen,
  Roger Grosse, Sam McCandlish, Jared Kaplan, Dario Amodei, Martin Wattenberg,
  and Christopher Olah.
\newblock Toy models of superposition, 2022.
\newblock URL \url{https://transformer-circuits.pub/2022/toy_model/index.html}.

\bibitem[Gekhman et~al.(2025)Gekhman, Ben~David, Orgad, Ofek, Belinkov,
  Szpektor, Herzig, and Reichart]{gekhman2025insideout}
Zorik Gekhman, Eyal Ben~David, Hadas Orgad, Eran Ofek, Yonatan Belinkov, Idan
  Szpektor, Jonathan Herzig, and Roi Reichart.
\newblock Inside-out: Hidden factual knowledge in {LLMs}.
\newblock In \emph{Conference on Language Modeling (COLM 2025)}, 2025.
\newblock \doi{10.48550/arXiv.2503.15299}.

\bibitem[Geva et~al.(2023)Geva, Bastings, Filippova, and
  Globerson]{geva2023dissecting}
Mor Geva, Jasmijn Bastings, Katja Filippova, and Amir Globerson.
\newblock Dissecting recall of factual associations in auto-regressive language
  models.
\newblock In \emph{Proceedings of the 2023 Conference on Empirical Methods in
  Natural Language Processing (EMNLP 2023)}, 2023.
\newblock \doi{10.48550/arXiv.2304.14767}.

\bibitem[Ghandeharioun et~al.(2024)Ghandeharioun, Caciularu, Pearce, Dixon, and
  Geva]{ghandeharioun2024patchscopes}
Asma Ghandeharioun, Avi Caciularu, Adam Pearce, Lucas Dixon, and Mor Geva.
\newblock Patchscopes: A unifying framework for inspecting hidden
  representations of language models.
\newblock In \emph{Proceedings of the 41st International Conference on Machine
  Learning}, volume 235 of \emph{Proceedings of Machine Learning Research},
  pages 15466--15490. PMLR, 2024.
\newblock \doi{10.48550/arXiv.2401.06102}.
\newblock URL \url{https://proceedings.mlr.press/v235/ghandeharioun24a.html}.

\bibitem[Hannun et~al.(2023)Hannun, Digani, Katharopoulos, and
  Collobert]{hannun2023mlx}
Awni Hannun, Jagrit Digani, Angelos Katharopoulos, and Ronan Collobert.
\newblock {MLX}: Efficient and flexible machine learning on apple silicon,
  2023.
\newblock URL \url{https://github.com/ml-explore/mlx}.

\bibitem[Heimersheim and Nanda(2024)]{heimersheim2024activation}
Stefan Heimersheim and Neel Nanda.
\newblock How to use and interpret activation patching, 2024.

\bibitem[Hewitt and Liang(2019)]{hewitt2019designing}
John Hewitt and Percy Liang.
\newblock Designing and interpreting probes with control tasks.
\newblock In \emph{Proceedings of the 2019 Conference on Empirical Methods in
  Natural Language Processing and the 9th International Joint Conference on
  Natural Language Processing (EMNLP-IJCNLP)}, pages 2733--2743, Hong Kong,
  China, 2019. Association for Computational Linguistics.
\newblock \doi{10.18653/v1/D19-1275}.

\bibitem[Levelt et~al.(1999)Levelt, Roelofs, and Meyer]{levelt1999theory}
Willem J.~M. Levelt, Ardi Roelofs, and Antje~S. Meyer.
\newblock A theory of lexical access in speech production.
\newblock \emph{Behavioral and Brain Sciences}, 22\penalty0 (1):\penalty0
  1--38, 1999.
\newblock \doi{10.1017/S0140525X99001776}.

\bibitem[Meng et~al.(2022)Meng, Bau, Andonian, and Belinkov]{meng2022locating}
Kevin Meng, David Bau, Alex Andonian, and Yonatan Belinkov.
\newblock Locating and editing factual associations in {GPT}.
\newblock In \emph{Advances in Neural Information Processing Systems},
  volume~35, 2022.

\bibitem[Miller(1995)]{miller1995wordnet}
George~A. Miller.
\newblock {WordNet}: a lexical database for {English}.
\newblock \emph{Communications of the ACM}, 38\penalty0 (11):\penalty0 39--41,
  1995.
\newblock \doi{10.1145/219717.219748}.

\bibitem[Morand et~al.(2025)Morand, Mothe, and Piwowarski]{morand2025entities}
Victor Morand, Josiane Mothe, and Benjamin Piwowarski.
\newblock On the representations of entities in auto-regressive large language
  models.
\newblock In \emph{Proceedings of the BlackBoxNLP Workshop at EMNLP 2025},
  2025.
\newblock \doi{10.48550/arXiv.2510.09421}.

\bibitem[Schaeffer et~al.(2023)Schaeffer, Miranda, and
  Koyejo]{schaeffer2023mirage}
Rylan Schaeffer, Brando Miranda, and Sanmi Koyejo.
\newblock Are emergent abilities of large language models a mirage?
\newblock In \emph{Advances in Neural Information Processing Systems},
  volume~36, 2023.
\newblock \doi{10.48550/arXiv.2304.15004}.

\bibitem[Vig et~al.(2020)Vig, Gehrmann, Belinkov, Qian, Nevo, Singer, and
  Shieber]{vig2020causal}
Jesse Vig, Sebastian Gehrmann, Yonatan Belinkov, Sharon Qian, Daniel Nevo,
  Yaron Singer, and Stuart Shieber.
\newblock Investigating gender bias in language models using causal mediation
  analysis.
\newblock In \emph{Advances in Neural Information Processing Systems},
  volume~33, 2020.
\newblock URL
  \url{https://proceedings.neurips.cc/paper/2020/hash/92650b2e92217715fe312e6fa7b90d82-Abstract.html}.

\bibitem[Wei et~al.(2022)Wei, Tay, Bommasani, Raffel, Zoph, Borgeaud, Yogatama,
  Bosma, Zhou, Metzler, Chi, Hashimoto, Vinyals, Liang, Dean, and
  Fedus]{wei2022emergent}
Jason Wei, Yi~Tay, Rishi Bommasani, Colin Raffel, Barret Zoph, Sebastian
  Borgeaud, Dani Yogatama, Maarten Bosma, Denny Zhou, Donald Metzler, Ed~H.
  Chi, Tatsunori Hashimoto, Oriol Vinyals, Percy Liang, Jeff Dean, and William
  Fedus.
\newblock Emergent abilities of large language models.
\newblock \emph{Transactions on Machine Learning Research}, 2022.
\newblock \doi{10.48550/arXiv.2206.07682}.

\bibitem[Xu et~al.(2024)Xu, Zhang, Zhang, Qian, and Huang]{xu2024tip}
Ningyu Xu, Qi~Zhang, Menghan Zhang, Peng Qian, and Xuanjing Huang.
\newblock On the tip of the tongue: Analyzing conceptual representation in
  large language models with reverse-dictionary probe, 2024.

\bibitem[Xu et~al.(2025)Xu, Zhang, Du, Luo, Qiu, Huang, and
  Zhang]{xu2025revealing}
Ningyu Xu, Qi~Zhang, Chao Du, Qiang Luo, Xipeng Qiu, Xuanjing Huang, and
  Menghan Zhang.
\newblock Revealing emergent human-like conceptual representations from
  language prediction.
\newblock \emph{Proceedings of the National Academy of Sciences}, 122\penalty0
  (44):\penalty0 e2512514122, 2025.
\newblock \doi{10.1073/pnas.2512514122}.

\bibitem[Yang et~al.(2024{\natexlab{a}})]{yang2024qwen2}
An~Yang et~al.
\newblock Qwen2 technical report, 2024{\natexlab{a}}.

\bibitem[Yang et~al.(2024{\natexlab{b}})Yang, Gribovskaya, Kassner, Geva, and
  Riedel]{yang2024latently}
Sohee Yang, Elena Gribovskaya, Nora Kassner, Mor Geva, and Sebastian Riedel.
\newblock Do large language models latently perform multi-hop reasoning?
\newblock In \emph{Proceedings of the 62nd Annual Meeting of the Association
  for Computational Linguistics (ACL 2024)}, 2024{\natexlab{b}}.
\newblock \doi{10.48550/arXiv.2402.16837}.

\bibitem[Yuchi et~al.(2026)Yuchi, Du, and Eisner]{yuchi2026numbers}
Fengting Yuchi, Li~Du, and Jason Eisner.
\newblock {LLMs} know more about numbers than they can say.
\newblock In \emph{Proceedings of the Conference of the European Chapter of the
  Association for Computational Linguistics (EACL 2026)}, 2026.

\bibitem[Zhang and Nanda(2024)]{zhang2024best}
Fred Zhang and Neel Nanda.
\newblock Towards best practices of activation patching in language models:
  Metrics and methods.
\newblock In \emph{The Twelfth International Conference on Learning
  Representations (ICLR 2024)}, 2024.
\newblock \doi{10.48550/arXiv.2309.16042}.

\end{thebibliography}

\clearpage
\appendix
\section{Prompt templates of set M}\label{app:prompts}
\begin{center}
\begin{minipage}{0.95\textwidth}
\footnotesize
\begin{verbatim}
FEWSHOT_M = "Description: a net or cloth hung between two posts
              or trees to lie or sleep in.\n
              This object is called hammock.\n\n"
            "Description: a pot with a spout and a lid, used for
              brewing and pouring tea.\n
              This object is called teapot.\n\n"
            "Description: an instrument a doctor puts on the chest
              to listen to the heart and lungs.\n
              This object is called stethoscope.\n\n"
# (line breaks above are for layout only; each string is one line)
NAMING_M  = FEWSHOT_M + "Description: {d}\nThis object is called"
READ_M    = FEWSHOT_M + "Description: {d}"      # H2' only
PATCHSCOPES target = "cat -> cat\n1135 -> 1135\nhello -> hello\n? ->"
\end{verbatim}
\end{minipage}
\end{center}

\providecommand{\FloatBarrier}{}
\section{Deviations from the pre-registration}\label{app:deviations}
\begin{table}[!h]
\centering
\caption{Deviations from the pre-registration and implementation details.}\label{tab:deviations}
\footnotesize
\begin{tabular}{@{}p{0.30\textwidth}p{0.64\textwidth}@{}}
\toprule
Item & Deviation and handling \\
\midrule
H2 $\to$ H2$'$ & Stop rule triggered; A versus C replaces A versus B. Decided after group sizes were known, before any set-M measurement. \\
H2$'$ test form & Plan v2.1 specified a TOST equivalence test; the pre-registration v1.2 replaced it by a non-inferiority test plus a separate deficit test (before measurement). \\
Pre-registration status & Internal SHA-256-frozen document, not an external registry entry. \\
A versus C on H1 measures & Pre-registration \S5(b) (manipulation check); omitted in the first draft, added in Section~\ref{sec:res-posthoc}. \\
Post-hoc analyses & Added after internal review, not pre-registered (Section~\ref{sec:posthoc}). \\
H2$'$ layer range & Fixed at 40--75\,\% depth instead of ``as in H1'', to avoid dependence on H1 results. \\
READ\_M & H2$'$ reads states at description tokens, not at ``called'', making the measure name-free. \\
Group C & Restricted to description-selection success, for symmetry with A. \\
CKA & Dropped; cosine similarity with CSLS throughout. \\
Tuned lens & Replaced by Patchscopes; shuffled item--description model realised as control (b). \\
Probes, reverse patch & Exploratory because H3 was dropped; not run. \\
Phi-3 precision & 4-bit weights (fp16 does not fit in 8\,GB); group formation ran in fp16 and was not repeated. \\
Qwen in H2$'$ & Descriptive only (15 pairs). \\
D-M4-1: control (f) & Injecting the name's own embedding behaves as a positive control, so gain$_f$ is negative by construction at later layers. (f) was removed from the (c) criterion and reported separately; the literal rule including (f) is also reported with its verdict. Fixed after 6 Qwen smoke-test items. \\
D-M4-2: Phi-3 activations & The cache check failed in fp16 (maximum difference 0.177 against tolerance 0.01), caused by shape-dependent fp16 rounding in quantised kernels. Activations were set to fp32 (weights stay 4-bit); back-end verification was repeated (maximum difference 0.0) and the cache check passed (maximum difference $1.7\times10^{-4}$). \\
KV-cache prefix & Not pre-registered. The shared few-shot prefix is computed once and reused. Maximum difference to the uncached path: 0.00033 (Qwen), 0.0066 (SmolLM2, close to the 0.01 tolerance), 0.00017 (Phi-3). \\
Control-name pool & Where $\pm1$ token yields fewer than 30 names (one 5-token name in SmolLM2), the window is widened until 30 are reached. \\
Process restarts & Phi-3 runs were restarted every 25 items and buffer caches limited, because of memory growth only; no numerical effect. \\
\bottomrule
\end{tabular}
\end{table}
\FloatBarrier

\section{Patching gains at fixed depths}\label{app:depth}
\begin{table}[ht]
\centering
\footnotesize
\caption{Multi-position patching gains (nats) at fixed depths, item type i. b: other descriptions; c: hard-negative description. Layer in parentheses. Values for types ii and iii are in the numbers file. $q$: BH-adjusted; all $q=0.0003$ unless stated.}
\label{tab:depth}
\begin{tabular}{llrrr}
\hline
Model & Control & 50\,\% depth & 62.5\,\% depth & 75\,\% depth \\
\hline
Qwen    & b & 7.54 (12)  & 6.65 (15)  & 1.99 (18) \\
        & c & 1.34 & 1.22 & 0.64 \\
SmolLM2 & b & 13.82 (12) & 10.64 (15) & 3.93 (18) \\
        & c & 3.05 & 2.32 & 0.85 \\
Phi-3   & b & 16.61 (16) & 7.59 (20)  & 1.11 (24) \\
        & c & 3.86 & 1.62 & 0.03 ($q=0.4128$) \\
\hline
\end{tabular}
\end{table}

\end{document}